# A New Gene Selection Algorithm using Fuzzy-Rough Set Theory for Tumor Classification

**Seyedeh Faezeh Farahbakhshian\*, Milad Taleby Ahvanooey\*\***

*\* School of Electrical and Computer Engineering, Shiraz University, Shiraz, P.O. Box 71935, Iran.*
*\*\* School of Information Management, Nanjing University, Nanjing, P.O. Box 210008 P.R, China.*
*Corresponding Authors: (E-mail: faezeh.farahbakhsh@gmail.com, M.Taleby@ieee.org)*

**Abstract:** In statistics and machine learning, feature selection is the process of picking a subset of relevant attributes for utilizing in a predictive model. Recently, rough set-based feature selection techniques, that employ feature dependency to perform selection process, have been drawn attention. Classification of tumors based on gene expression is utilized to diagnose proper treatment and prognosis of the disease in bioinformatics applications. Microarray gene expression data includes superfluous feature genes of high dimensionality and smaller training instances. Since exact supervised classification of gene expression instances in such high-dimensional problems is very complex, the selection of appropriate genes is a crucial task for tumor classification. In this study, we present a new technique for gene selection using a discernibility matrix of fuzzy-rough sets. The proposed technique takes into account the similarity of those instances that have the same and different class labels to improve the gene selection results, while the state-of-the art previous approaches only address the similarity of instances with different class labels. To meet that requirement, we extend the Johnson reducer technique into the fuzzy case. Experimental results demonstrate that this technique provides better efficiency compared to the state-of-the-art approaches.

## 1. INTRODUCTION

Nowadays, "cancer" is one of the deadliest diseases around the world. According to a report evaluated by WHO in 2015, there has been a huge growth "8.8" million deaths caused by cancer. Consequently, it will be increased every year, if there is no way to diagnose such kind of diseases earlier (Aydadenta and Adiwijaya, 2018). In general, traditional ways of diagnosis are error prone and time consuming as they rely on individual arbitration. Therefore, the machine learning and pattern recognition algorithms could enhance the process of diagnosis and treatment of the illness in the field of health informatics and biomedical (Arunkumar and Ramakrishnan, 2018). A tumor is an abnormal growth of cells that is known as one of the systematic biology diseases. It can grow and behave differently than the mechanism expansion is not entirely identified yet. Also, a benign tumor is not cancer, but it could be benign pre-malignant or malignant. In general, there exist various types of tumors and diversity of names for them. The assigned names generally reflect their formation and the kind of tissue that they become manifest in each one. In other words, a tumor is a type of swelling or mass and does not pose a health threat. Due to the tumor medication of patients are in the later stage of cancer diagnosis, they are often not well-treated. Therefore, physicians concur that initial diagnosis of tumor is a crucial benefit for the prosperous therapies (Dai and Xu, 2013).

Over past two decades, gene expression based on the molecular assessment of the tumor has drawn attention from physician community due to providing higher detection accuracy and early tumor diagnosis. Basically, gene expression profiling is the measurement of the activity of thousands of genes at a once, to generate a global observation of cellular function in the molecular biology filed. For instance, these forms could identify the activity between cells which actively distribute or express how the cells react to a specific treatment. Nevertheless, there is a huge number of effective genes in the gene expression datasets, since simply some of them are crucial for classification. Therefore, an efficient algorithm for identifying proper genes becomes a vital problem for tumor detection (Zhang et al., 2005; Xu et al., 2011; Dai and Xu, 2013).

There have been introduced many approaches for gene selection over last two decades. One of the efficient methods to solve this issue is based on the rough set theory (RST) that has been successfully employed as the pre-processor of a dataset. The RST is a powerful mathematical method for dealing with ambiguous data, and it works according to crisp equivalence relations and equivalence classes (Pawlak, 1982; Khan et al., 2001). The primary idea of RST is to reduce the redundancy of data using attribute reduction in a dataset which only considers the data alone, without requiring extra information while preserving the ability of classification. It has attracted the attention of several scientists who have investigated the RST applications and theories in different scientific research domains (Khan et al., 2002; Thangavel and Pethalakshmi, 2009; Jensen and Shen, 2009; Kaneiwa, 2011;

Chen and Cheng, 2012; Dai and Xu, 2012; Jensen et al., 2014; Qian et al., 2015). However, the classical RST could only be applied to specific feature reduction and knowledge discovery. To deal with numerical and accurate data (or a mix of both), as gene expression data which are always continuous, the fuzzy rough set theory (FRST) was first introduced in (Dubois and Prade, 1990), that merges rough and fuzzy sets. In FRST, a fuzzy similarity relation is employed to determine the rate of similarity between two objects instead of the equivalence relation applied in the crisp rough sets (Wu et al., 2003; Jensen and Shen, 2009; Jensen et al., 2014).

Rough sets permit the generation from clauses of attributes in conjunctive normal form (CNF). In the case that we assign precision values to all items occurring in the clauses, the equation is satisfied, hence, the set of remaining features is a valid subset of the data. The task is to obtain the minimum number of features such that the CNF equation is satisfied. The problem of finding the smallest subset of features (called as reduct) utilizing the RST could be expressed as propositional satisfiability (SAT) difficulty which is NP-complete. An efficient way of solving this problem is to employ Davis–Putnam-Logemann–Loveland (DPLL) algorithm. However, it guarantees to obtain the minimal subset of features on big datasets such as tumors, but it suffers from high complexity (Øhrn, 1999; Wang et al., 2008; Jensen wt al., 2014). The computational cost of this technique at worst case is $O(2^n)$, where n is the number of features. So, we can utilize a heuristic technique such as Johnson Reducer (JR) algorithm for solving the CNF problem. Technically, it does not guarantee the minimality of features, but its result is typically close to minimal.

In this research, we introduce a new technique for improving gene selection based on discernibility matrix of fuzzy-rough sets. Our technique considers both the similarity of instances with the different class labels and the similarity of samples with the identical class labels. We extend the JR method in the fuzzy case to find the reduction of genes.

The remainder of this article is organized as follows. Section 2 presents the background on information systems and fuzzy-rough sets. Section 3 introduces the proposed gene selection technique based on fuzzy discernibility matrix (FDM) and JR algorithm. Section 4 expresses the experiments of the proposed technique conducted on five gene expression tumor datasets and compares it with the-state-of-the-art approaches. Finally, section 5 draws the concluding remarks.

## 2. ROUGH SET THEORY

In theoretical computer science, a rough set is a formal approximation of a crisp set regarding a pair of sets that present the lower (LA) and the upper approximation (UA) of the original set. The LA is the set of objects which belong to the custom subset, and the UA is the set of objects which possibly relate to the subset (Khan et al., 2001). For finding reducts based on the RST, there are two main approaches: dependency degree-based and discernibility matrix (DM) based algorithms.

*2.1 Decision systems*

Let $\mathbb{A} = \{\mathbb{U}, \mathbb{C} \cup \mathbb{D}\}$ is a decision system where $\mathbb{U}$ is a set of objects, and $\mathbb{C} \cup \mathbb{D}$ is the set of provisional and decision features. Herein, $\mathbb{D}$ is the set of class or decision labels. Also, for each $d \in D$, there is a function $d: \mathbb{U} \rightarrow \{0,1\}$ which shows each object $a \in \mathbb{U}$ belongs to class $d$ ($d(a) = 1\ or\ 0$). An instance of decision system illustrated in Table 1 in which $\mathbb{U} = \{x_0, x_1, \ldots, x_7\}$ is the set of objects and $\{a, b, c, d, q\}$ is the set of features. Moreover, the conditional features are $\mathbb{C} = \{a, b, c, d\}$ and the decision feature is $\mathbb{D} = \{q\}$.

**Table 1. A synthetic dataset with nominal values.**

| $x \in \mathbb{U}$ | a | b | c | d | q |
|---|---|---|---|---|---|
| $x_0$ | 2 | 0 | 1 | 1 | 0 |
| $x_1$ | 0 | 2 | 2 | 2 | 1 |
| $x_2$ | 1 | 0 | 0 | 2 | 2 |
| $x_3$ | 2 | 2 | 0 | 1 | 1 |
| $x_4$ | 2 | 0 | 1 | 0 | 2 |
| $x_5$ | 1 | 1 | 0 | 2 | 2 |
| $x_6$ | 1 | 2 | 2 | 2 | 1 |
| $x_7$ | 0 | 2 | 2 | 0 | 2 |

*2.2 Discernibility matrix (DM)*

Recently, several RST based feature selection methods have employed the discernibility matrices for obtaining the reducts (Jensen and Shen, 2009; Jensen et al., 2014). In a decision table $(\mathbb{U}, \mathbb{C} \cup \mathbb{D})$, a DM is a symmetric $|\mathbb{U}| \times |\mathbb{U}|$ matrix with entries that can be expressed as (Skowron and Rauszer, 1992):

$$C_{ij} = \{a \in \mathbb{C} \mid a(x_i) \neq a(x_j)\},$$
$$i, j = 1, \ldots, |\mathbb{U}| \quad (1)$$

Each $C_{ij}$ is a clause including features that discriminate between objects $x_i$ and $x_j$. One of the effective techniques for finding reducts is the use of DM. This process only considers the discernibility of objects that have the difference decision features. According to the values in Table 1, the DM can be depicted as matrix $C$. Here, objects $x_0$ and $x_1$ are different in all features. Also, some elements in objects $x_1$ and $x_3$ are different, the clause is empty because the corresponding decision feature $q$ is identical.

To obtain the minimum set of features (reducts) which discriminate between the objects, we can use the conjunctive (or disjunctive) normal form of clauses that named as discernibility functions (Pawlak and Skowron, 2007). Discernibility function $f_D$ for $m$ features $\{a_1, \ldots, a_m\}$ defined as:

$$f_D(a_1, \ldots, a_m) = \{\wedge \{\vee C_{ij}\} \mid 1 \leq j \leq i \leq |\mathbb{U}|, C_{ij} \neq \emptyset\} \quad (2)$$

where ∧ and ∨ indicate the logical operators AND and OR, respectively, and $C_{ij}$ is a clause, as defined in equation (1).

As depicted in Table 1, the discernibility function (after removing the duplicates) is $f_D(a, b, c, d) = (a \vee b \vee c \vee d) \wedge (a \vee c \vee d) \wedge (b \vee c) \wedge (d) \wedge (a \vee b \vee c) \wedge (a \vee b \vee d) \wedge (b \vee c \vee d) \wedge (a \vee d)$. This function can be still simplified by taking away the clauses that are subsumed with other values: $f_D(a, b, c, d) = (b \vee c) \wedge (d)$. Therefore, the minimum reducts are {b,d} or {c,d}. However, this method ensures to find all reducts, but it has very high complexity for even datasets with medium sizes.

$$C = \begin{pmatrix} \{\} & \{a,b,c,d\} & \{a,c,d\} & \{b,c\} & \{d\} & \{a,b,c,d\} & \{a,b,c,d\} & \{a,b,c,d\} \\ \{a,b,c,d\} & \{\} & \{a,b,c\} & \{\} & \{a,b,c,d\} & \{a,b,c\} & \{\} & \{d\} \\ \{a,c,d\} & \{a,b,c\} & \{\} & \{a,b,d\} & \{\} & \{\} & \{b,c\} & \{\} \\ \{b,c\} & \{\} & \{a,b,d\} & \{\} & \{b,c,d\} & \{a,b,d\} & \{\} & \{a,c,d\} \\ \{d\} & \{a,b,c,d\} & \{\} & \{b,c,d\} & \{\} & \{\} & \{a,b,c,d\} & \{\} \\ \{a,b,c,d\} & \{a,b,c\} & \{\} & \{a,b,d\} & \{\} & \{\} & \{b,c\} & \{\} \\ \{a,b,c,d\} & \{\} & \{b,c\} & \{\} & \{a,b,c,d\} & \{b,c\} & \{\} & \{a,d\} \\ \{a,b,c,d\} & \{d\} & \{\} & \{a,c,d\} & \{\} & \{\} & \{a,d\} & \{\} \end{pmatrix}$$

There is another solution to this problem based on the JR method, which can obtain the reducts without guaranteeing the minimality (Øhrn, 1999). Nevertheless, its size is close to the minimal. The recursive form of the JRA algorithm is shown in algorithm 1.

---
**Algorithm 1:** Johnson reducer algorithm.
JRA $(F, R)$
$F$: discernibility function of current clauses.
$R$: set of selected features.
1. **if** ($F$ is empty)
2.     **output:** current assignment, $R$;
3.     **return:** satisfiable;
4. **else if** ($F$ includes a unit clause)
5.     $(F', R') \leftarrow$ unitPropagate$(F)$;
6.     **return** JRA$(F', R')$;
7. **else**
8.     $x \leftarrow$ selectLiteral $(F)$
9.     $(F', R') \leftarrow$ Propagate$(F, x)$
10.    **return** JRA$(F', R')$

---

In the JRA, the clauses in the discernibility function consist of a single literal called unit clauses and can be satisfied. Since the deletion of a literal step could generate some new unit clauses, the unitPropagate phase repeats until there is no any unit clause in the discernibility function. The selection of unit literals from unit clauses and also updating the clause list are repeated until there is not any unit clause. Then, each conditional feature that appears in the discernibility function is evaluated according to the heuristic measure (in line 8). It typically considers the literal to compute the number of times that a feature appears within clauses. The features that appear more frequently are supposed to be more significant. So, each attribute with the highest heuristic value in discernibility function is added to the reduct candidate and all other clauses containing this feature will be removed. When all clauses removed, this process will be terminated. This technique returns the set $R$ which guarantees to be a reduct if all clauses in the function are satisfied (Jensen et al., 2014).

*2.3. Fuzzy discernibility matrix*

In theory, the RST feature selection can only deal with the data sets containing symbolic features such as tumor datasets in which the discretization process is essential. However, this process leads to information loss. To work with a real dataset that consist of real values, a combination of fuzzy and rough sets called fuzzy-rough sets, is presented in (Dubois and Prade, 1990), which can provide better feature selection rate.

A synthetic dataset with real-valued features is depicted in Table 2. In this table, the assumed objects are: $\mathbb{U} = \{x_0, \ldots, x_5\}$, $\mathbb{C} = \{a, b, c\}$ and $\mathbb{D} = \{q\}$.

The FDM is the extent of the crisp rough set for finding fuzzy-rough reducts. The first step is provided in (Tsang et al., 2005; Chen et al., 2012) where a crisp DM is created using "threshold" and converting the real-value features to the crisp-value for fuzzy-rough feature selection. This is in contrast to Rough's theory, that does not need any extra information except for finding reducts.

Table 2. Another synthetic dataset with real-valued features.

| $x \in \mathbb{U}$ | a | b | c | q |
|---|---|---|---|---|
| $x_0$ | -0.3 | -0.3 | -0.5 | 0 |
| $x_1$ | -0.2 | -0.5 | -0.3 | 1 |
| $x_2$ | -0.3 | -0.4 | 0.1 | 0 |
| $x_3$ | -0.4 | 0.1 | 0.1 | 0 |
| $x_4$ | 0.2 | -0.1 | -0.1 | 1 |
| $x_5$ | -0.3 | -0.6 | 0.1 | 1 |

In the FDM, Entries of the discernibility matrix is a fuzzy set wherein each feature belongs to a certain degree. In other words, for each feature $a$ in the fuzzy clause $C_{ij}$, the fuzzy discernibility measure is as follows (Jensen et al., 2014):

$$\mu_{C_{ij}}(a) = N\left(\mu_{R_a}(i,j)\right) \quad (3)$$

where $N(.)$ defines fuzzy negation and $\mu_{R_a}(i,j)$ is the fuzzy similarity (i.e., similarity rate) of objects $x_i$ and $x_j$ in $\mathbb{U}$ according to feature $a$. To meet this requirement, a fuzzy similarity measure can be expressed as follows:

$$\mu_{R_a}(i,j) = \begin{cases} 1 - 4 * \dfrac{|a(x_i) - a(x_j)|}{|a_{max} - a_{min}|}, & \text{if } \dfrac{|a(x_i) - a(x_j)|}{|a_{max} - a_{min}|} \leq 0.25 \\ 0, & \text{otherwise} \end{cases} \quad (4)$$

and

$$\mu_{R_a}(i,j) = max\left(min\left(\dfrac{a(x_j) - (a(x_i) - \sigma_a)}{\sigma_a}, \dfrac{(a(x_i) + \sigma_a) - a(x_j)}{\sigma_a}\right), 0\right) \quad (5)$$

Where $\sigma_a$ is the standard deviation of feature $'a'$ over all objects $x_i$, $i = 1, ..., |\mathbb{U}|$. Therefore, $\mu_{C_{ij}}(a)$ is a fuzzy discernibility measure. In the crisp case, if $\mu_{C_{ij}}(a) = 1$, two objects $x_i$ and $x_j$ are discernible for feature $a$; if $\mu_{C_{ij}}(a) = 0$, these objects are equal. In fuzzy cases, where $\mu_{C_{ij}}(a) \epsilon (0,1)$, the objects with a certain degree are discernible. In theory, each entry in the FDM is a set of features and their memberships:

$$C_{ij} = \{a_k \mid a \epsilon \mathbb{C}, \quad k = N\left(\mu_{R_a}(i,j)\right)\}, \quad i,j = 1, ..., |\mathbb{U}| \quad (6)$$

For example, the clause $C_{ij}$ can be: $\{a_{0.8}, b_{0.6}, c_{0.1}, d_{0.0}\}$ that $\mu_{C_{ij}}(a) = 0.8$, $\mu_{C_{ij}}(b) = 0.6$, etc. In the crisp discernibility matrix, these values are 0 or 1. These scores represent the significance of each feature in the clause that discriminates between two factors $x_i$ and $x_j$.

As in the crisp discernibility function, only clauses with dissimilar features are considered in the fuzzy case that can be extended as follows (Jensen et al., 2014):

$$f_D(a_1, ..., a_m) = \left\{ \wedge \{\vee C_{ij} \leftarrow q_{\mu_{C_{ij}}(q)}\} | 1 \leq j \leq i \leq |\mathbb{U}| \right\} \quad (7)$$

where $q$ is the assigned feature. If $\mu_{C_{ij}}(q) = 1$ then clause $C_{ij}$ is maximally discernible for objects $x_i$ and $x_j$.

For solving and simplifying the discernibility function in equation (7), a fuzzy extension of JR algorithm (FJRA) is presented. To meet this requirement, the recursive JR algorithm is employed for finding the fuzzy-rough reduct as depicted in algorithm 2. First, it inputs an empty set of reducts, $R$. Then, it uses the unit clauses in the discernibility function (list of clauses). Therefore, it employs the unit propagation to the current function (in line 5) and removes the selected literal from it until there is no unit clause in the discernibility function. Then, the FJRA uses a literal which is selected from the current clauses (see line 8). The selection process of proper literals is effective due to choosing different heuristics may produce different size reducts. Thus, it could enhance the efficiency of the selected features. One of the effective heuristic methods is the selection of features that have the non-zero fuzzy discernibility in most of the current clauses. After choosing a proper literal, all clauses which are maximally satisfied, will be removed. In addition, the rest of clauses are added to the new set of clauses, $F'$ (in line 13). For each clause $C_{ij}$, the maximum satisfiability degree can be calculated as:

$$maxSAT_{ij} = S_{a \epsilon \mathbb{C}} \{\mu_{C_{ij}}(a)\} \quad (8)$$

here $S$ is a t-conorm operator and the maximal amount that clause $C_{ij}$ may be satisfied. Finally, the FJRA is terminated by removing all clauses from discernibility function and returns the reduct, $R$.

To apply the FJRA algorithm on clauses list, we perform the simplification phase as a preprocessing step. This step on the crisp discernibility function involves removing clauses that are supersets of others. In the fuzzy case, the simplification can still be performed by removing redundant clauses that satisfied by others in the certain degree. Moreover, according to properties of fuzzy implication, those redundant clauses which have zero decision component, can also be removed (Jensen et al., 2014). Therefore, the simplification degree in the crisp case can be extended as:

$$S(C_{ij}, C_{kl}) = \frac{\sum_{a \epsilon \mathbb{C}} T(\mu_{C_{ij}}(a), \mu_{C_{kl}}(a))}{\sum_{a \epsilon \mathbb{C}} \mu_{C_{ij}}(a)} \quad (9)$$

**Algorithm 2:** Fuzzy Johnson reducer algorithm (Jensen et al., 2014)

FJRA $(F, R)$
$F$: discernibility function of current clauses.
$R$: set of selected features.
1.  if ($F$ is empty)
2.      output: current assignment, $R$;
3.      return: satisfiable;
4.  else if($F$ includes a unit clause)
5.      $(F', R') \leftarrow$ unitPropagate$(F)$;
6.      return FJRA$(F', R')$;
7.  else
8.      $x \leftarrow$ selectLiteral $(F)$;
9.      $R' \leftarrow R \cup \{x\}$;
10.     $F' \leftarrow \emptyset$ ;
11.     for each $f \epsilon F$
12.         if ( !isSatisfied$(f)$ )
13.             $F' \leftarrow F' \cup \{f\}$;
14.     end for
15.     return FJRA$(F', R')$

If $S(C_{ij}, C_{kl}) = 1$ then clause $C_{kl}$ can be removed because this is subsumed by clause $C_{ij}$.

This reduction process is effective for removing the redundant clauses, but computationally expensive. It needs each clause in the list to be compared with others. In the worst case, $c = (n^2 - n)/2$ clauses are produced initially which indicates the number of comparisons. However, it can be diminished by combining the reduction process into the generating DM clauses.

As an example, we considered the variables and values in Table 2 as a decision dataset with real-valued features. The fuzzy similarity matrix of objects for each feature can be obtained using equation (4) as:

$$\mu_{R_a} = \begin{pmatrix} 1 & 0.33 & 1 & 0.33 & 0 & 1 \\ 0.33 & 1 & 0.33 & 0 & 0 & 0.33 \\ 1 & 0.33 & 1 & 0.33 & 0 & 1 \\ 0.33 & 0 & 0.33 & 1 & 0 & 0.33 \\ 0 & 0 & 0 & 0 & 1 & 0 \\ 1 & 0.33 & 1 & 0.33 & 0 & 1 \end{pmatrix},$$

$$\mu_{R_b} = \begin{pmatrix} 1 & 0.43 & 1 & 0 & 0 & 0 \\ 0.43 & 1 & 0.43 & 0 & 0 & 0.43 \\ 1 & 0.43 & 1 & 0 & 0 & 0 \\ 0 & 0 & 0 & 1 & 0 & 0 \\ 0 & 0 & 0 & 0 & 1 & 0 \\ 0 & 0.43 & 0 & 0 & 0 & 1 \end{pmatrix},$$

$$\mu_{R_c} = \begin{pmatrix} 1 & 0 & 0 & 0 & 0 & 0 \\ 0 & 1 & 0 & 0 & 0 & 0 \\ 0 & 0 & 1 & 1 & 0 & 1 \\ 0 & 0 & 1 & 1 & 0 & 1 \\ 0 & 0 & 0 & 0 & 1 & 0 \\ 0 & 0 & 1 & 1 & 0 & 1 \end{pmatrix},$$

$$\mu_{R_d} = \begin{pmatrix} 1 & 0 & 1 & 1 & 0 & 0 \\ 0 & 1 & 0 & 0 & 1 & 1 \\ 1 & 0 & 1 & 1 & 0 & 0 \\ 1 & 0 & 1 & 1 & 0 & 0 \\ 0 & 1 & 0 & 0 & 1 & 1 \\ 0 & 1 & 0 & 0 & 1 & 1 \end{pmatrix}.$$

Next, the fuzzy discernibility matrices are constructed by using the equation (3). For objects $x_1$ and $x_2$, the fuzzy clause is $C_{12} = \{a_{0.67} \vee b_{1.00} \vee c_{1.00} \leftarrow q_{1.0}\}$. In the case that the fuzzy discernibility of objects $x_1$ and $x_2$ for feature $a$ is 0.67, it shows that the objects $x_1$ and $x_2$ are partly discernible for this feature. These objects are fully discernible according to the decision feature as depicted in $q_{1.0}$. In this step, the set of generated clauses can be expressed as:

$C_{12} = \{a_{0.67} \vee b_{1.00} \vee c_{1.00} \leftarrow q_{1.0}\}$,
$C_{13} = \{a_{0.00} \vee b_{0.57} \vee c_{1.00} \leftarrow q_{0.0}\}$,
$C_{14} = \{a_{0.67} \vee b_{1.00} \vee c_{1.00} \leftarrow q_{0.0}\}$,
$C_{15} = \{a_{1.00} \vee b_{1.00} \vee c_{1.00} \leftarrow q_{1.0}\}$,
$C_{16} = \{a_{0.00} \vee b_{1.00} \vee c_{1.00} \leftarrow q_{1.0}\}$,
$C_{23} = \{a_{0.67} \vee b_{0.57} \vee c_{1.00} \leftarrow q_{1.0}\}$,
$C_{24} = \{a_{1.00} \vee b_{1.00} \vee c_{1.00} \leftarrow q_{1.0}\}$,
$C_{25} = \{a_{1.00} \vee b_{1.00} \vee c_{1.00} \leftarrow q_{0.0}\}$,
$C_{26} = \{a_{0.67} \vee b_{0.57} \vee c_{1.00} \leftarrow q_{0.0}\}$,
$C_{34} = \{a_{0.67} \vee b_{1.00} \vee c_{0.00} \leftarrow q_{0.0}\}$,
$C_{35} = \{a_{1.00} \vee b_{1.00} \vee c_{1.00} \leftarrow q_{1.0}\}$,
$C_{36} = \{a_{0.00} \vee b_{1.00} \vee c_{0.00} \leftarrow q_{1.0}\}$,
$C_{45} = \{a_{1.00} \vee b_{1.00} \vee c_{1.00} \leftarrow q_{1.0}\}$,
$C_{46} = \{a_{0.67} \vee b_{1.00} \vee c_{0.00} \leftarrow q_{1.0}\}$,
$C_{56} = \{a_{1.00} \vee b_{1.00} \vee c_{1.00} \leftarrow q_{0.0}\}$.

According to the attributes of implications, all clauses with $q_{0.0}$ are eliminated without affecting the final reduct. Also, the identical clauses are redundant. So, the clause list is reduced to:

$C_{12} = \{a_{0.67} \vee b_{1.00} \vee c_{1.00} \leftarrow q_{1.0}\}$,
$C_{15} = \{a_{1.00} \vee b_{1.00} \vee c_{1.00} \leftarrow q_{1.0}\}$,
$C_{16} = \{a_{0.00} \vee b_{1.00} \vee c_{1.00} \leftarrow q_{1.0}\}$,
$C_{23} = \{a_{0.67} \vee b_{0.57} \vee c_{1.00} \leftarrow q_{1.0}\}$,
$C_{36} = \{a_{0.00} \vee b_{1.00} \vee c_{0.00} \leftarrow q_{1.0}\}$,
$C_{46} = \{a_{0.67} \vee b_{1.00} \vee c_{0.00} \leftarrow q_{1.0}\}$.

To apply the FJRA algorithm, first, we need to simplify the clauses list using equation (9). The simplest clause list is:

$C_{23} = \{a_{0.67} \vee b_{0.57} \vee c_{1.00} \leftarrow q_{1.0}\}$,
$C_{36} = \{a_{0.00} \vee b_{1.00} \vee c_{0.00} \leftarrow q_{1.0}\}$

Then, the FJRA algorithm is employed to discover the reducts. Where, there is only one unit clause in this list, the feature b is chosen as the first selected feature and this clause is satisfied. Since there is not any unit clause in remaining list, the next feature according to the highest value of fuzzy discernibility degree is selected. Then, the feature c is chosen and all clauses will be satisfied. Finally, the feature set $\{b, c\}$ is selected.

### 2.4. Related Work

In this subsection, we summarize some feature selection techniques that have been presented according to FRST recently. Also, we highlight their pros and cons in order to show their limitations and advantages.

R. Jensen and Q. Shen (Jensen and Shen, 2009) proposed three feature selection approaches based on fuzzy-rough and fuzzy similarity relation. The first technique works based on fuzzy lower approximations, which generates a new measurement of attribute dependency using similarity relation. The second technique utilizes the fuzzy boundary region information to train the search process of feature selection. It generates a fuzzy-rough reduct when this criterion minimized. The third technique extends the discernibility matrix concepts to the fuzzy manner, called FDM that assigns a certain degree to each entry of discernibility matrix. In practice, the FDM is computationally expensive with long run times (i.e., $O((n2 + n)/2)$).

R. Jensen et al. (Jensen et al., 2014) suggested a new technique based on the DPLL approach for finding the minimum subset of features based on rough and fuzzy-rough contexts, called JRA. Technically, it provides an extension of the DM to the fuzzy manner, and generates clauses by fuzzy similarity relation and SAT search method. In this work, the global minimum reduct for a particular dataset could be computed. This generalization extends the use of this technique for discrete and continuous values datasets. However, this technique is not computationally expensive but does not guarantee global minimality.

Dai et al. (Dai et al., 2017) proposed two algorithms for feature selection from the view of object pair in the area of fuzzy-rough sets called RMDPS and WRMDPS. The RMDPS is the summarization of Reduced Maximal Discernibility Pairs Selection and the WRMDPS is Weighted RMDPS. These mechanisms simply need to deal with part of the object pairs instead of all of the object pairs, that increases the efficiency of the feature selection rate.

Raza and Qamar (Raza and Qamar, 2016) presented a new heuristic-based technique called Incremental Dependency Class (IDCs) for calculating the dependency in the RST. The conventional technique for obtaining the dependency consists of computing the positive region. The merit of the IDCs is to avoid the calculation of this time-expensive task, which makes them applicable to the selection of features in the large datasets and ensures some optimality of the small feature subsets. However, it does not guarantee the finding of the best larger subset.

Das et al. (Das et al., 2018) introduced an attribute selection technique based on RST and genetic algorithm for extracting the optimum subset of attributes. This algorithm provides a more compact attribute subset using genetic algorithm and can select attributes in the static and dynamic environment. However, it has very high computational cost due to the evaluation of each gene needs building a predictive model.

Dara et al. (Dara, et al, 2017) provided a hybrid algorithm utilizing the RST and binary Particle Swarm Optimization (PSO), which can obtain the minimum subset of features for high dimensional cancer datasets. In this work, the researchers utilized a fast heuristic scheme to decrease domain features using the elimination of redundant attributes statistically. Moreover, they discretized datasets into a "binary table" called distinction table in RST. Practically, since this table is employed to assess and enhance the objectives functions, it leads to lose some useful information.

## 3. PROPOSED TECHNIQUE

In this section, we proposed a new technique to improve the gene selection results based on fuzzy-rough sets for biomedical applications. In recent researches (Jensen and Shen, 2009; Jensen et al., 2014), there exist two problems that we address them in this paper. First, the significance degree of each feature was computed by negating the fuzzy similarity of two objects, i.e., see the equation (3). Also, they did not discriminate between the objects with the same class labels and different labels. The second problem was the discernibility function expressed in (4), which neglects the impact of the objects with the same class labels (due to implication operator). Herein, we try to address mentioned problems by presenting our solutions. To solve the first problem, we define a new criterion to measure the fuzzy discernibility of genes in each clause in order to discriminate the objects with the same class labels from different labels. This criterion is defined as:

$$\mu_{C_{ij}}(a) = \begin{cases} N\left(\mu_{R_a}(i,j)\right), & \text{if } q(i) \neq q(j) \\ \mu_{R_a}(i,j), & \text{if } q(i) = q(j) \end{cases} \quad (10)$$

where $\mu_{C_{ij}}(a)$ is the significance degree of gene $a$ in clause $C_{ij}$ for objects $x_i$ and $x_j$ according to class label $q$. In the case that, two objects have different class labels, the significance degree is measured using the negation of their fuzzy similarity. Where, the similarity rate of two objects increases, the significance degree of this gene in the corresponding clause decreases. On the other hand, if two objects have the identical class labels, the significance degree is measured using their fuzzy similarity in order to enhance the significance of genes in clauses.

Additionally, to enhance the effectiveness of objects having the same class labels, we define a new discernibility function according to membership of genes. This new function is expressed as:

$$f_D(a_1, \dots, a_m) = \{\wedge \{\vee C_{ij} \leftarrow q_{1.0}\} | 1 \leq j \leq i \leq |\mathbb{U}|\} \quad (11)$$

The difference between functions in equation (11) and (7) is that the membership degree of $q$ in equation (11) is always 1.0. This is because of new membership degree in (10) which discriminates objects with the same class labels and different class labels. If two objects $x_i$ and $x_j$ have the identical class labels, $\mu_{R_q}(i,j) = 1$ and so $\mu_{C_{ij}}(q) = \mu_{R_q}(i,j) = 1$. On the other hand, if the class labels are different, $\mu_{R_q}(i,j) = 0$ and again $\mu_{C_{ij}}(q) = N(\mu_{R_q}(i,j)) = 1$. Thus always $\mu_{C_{ij}}(q) = 1$.

Therefore, the satisfied clauses using equation (11) could be more than (or at least equal to) those satisfied by equation (7). To find the reduct from the new clause list, the FJRA is used once again.

To obtain the fuzzy rough set reduct for the synthetic dataset in Table 2, first, the fuzzy similarity matrix of objects for each gene is calculated using equation (4). Then, the FDM is constructed based on equation (11). The set of clauses can be expressed:

$C_{12} = \{a_{0.67} \vee b_{1.00} \vee c_{1.00} \leftarrow q_{1.0}\}$,
$C_{13} = \{a_{1.00} \vee b_{0.43} \vee c_{0.00} \leftarrow q_{1.0}\}$,
$C_{14} = \{a_{0.33} \vee b_{0.00} \vee c_{0.00} \leftarrow q_{1.0}\}$,
$C_{15} = \{a_{1.00} \vee b_{1.00} \vee c_{1.00} \leftarrow q_{1.0}\}$,
$C_{16} = \{a_{0.00} \vee b_{1.00} \vee c_{1.00} \leftarrow q_{1.0}\}$,
$C_{23} = \{a_{0.67} \vee b_{0.57} \vee c_{1.00} \leftarrow q_{1.0}\}$,
$C_{24} = \{a_{1.00} \vee b_{1.00} \vee c_{1.00} \leftarrow q_{1.0}\}$,
$C_{25} = \{a_{0.00} \vee b_{0.00} \vee c_{0.00} \leftarrow q_{1.0}\}$,
$C_{26} = \{a_{0.33} \vee b_{0.43} \vee c_{0.00} \leftarrow q_{1.0}\}$,
$C_{34} = \{a_{0.33} \vee b_{0.00} \vee c_{1.00} \leftarrow q_{1.0}\}$,
$C_{35} = \{a_{1.00} \vee b_{1.00} \vee c_{1.00} \leftarrow q_{1.0}\}$,
$C_{36} = \{a_{0.00} \vee b_{1.00} \vee c_{0.00} \leftarrow q_{1.0}\}$,
$C_{45} = \{a_{1.00} \vee b_{1.00} \vee c_{1.00} \leftarrow q_{1.0}\}$,

$C_{46} = \{a_{0.67} \lor b_{1.00} \lor c_{0.00} \leftarrow q_{1.0}\}$,
$C_{56} = \{a_{0.00} \lor b_{0.00} \lor c_{0.00} \leftarrow q_{1.0}\}$.

Hence, the duplicate clauses and empty clauses (e.g., $C_{25}$) are removed as they can not be satisfied. So, the clause list is reduced to:

$C_{12} = \{a_{0.67} \lor b_{1.00} \lor c_{1.00} \leftarrow q_{1.0}\}$,
$C_{13} = \{a_{1.00} \lor b_{0.43} \lor c_{0.00} \leftarrow q_{1.0}\}$,
$C_{14} = \{a_{0.33} \lor b_{0.00} \lor c_{0.00} \leftarrow q_{1.0}\}$,
$C_{15} = \{a_{1.00} \lor b_{1.00} \lor c_{1.00} \leftarrow q_{1.0}\}$,
$C_{16} = \{a_{0.00} \lor b_{1.00} \lor c_{1.00} \leftarrow q_{1.0}\}$,
$C_{23} = \{a_{0.67} \lor b_{0.57} \lor c_{1.00} \leftarrow q_{1.0}\}$,
$C_{26} = \{a_{0.33} \lor b_{0.43} \lor c_{0.00} \leftarrow q_{1.0}\}$,
$C_{34} = \{a_{0.33} \lor b_{0.00} \lor c_{1.00} \leftarrow q_{1.0}\}$,
$C_{36} = \{a_{0.00} \lor b_{1.00} \lor c_{0.00} \leftarrow q_{1.0}\}$,
$C_{46} = \{a_{0.67} \lor b_{1.00} \lor c_{0.00} \leftarrow q_{1.0}\}$.

Using (9), the clause list is simplified to:

$C_{14} = \{a_{0.33} \lor b_{0.00} \lor c_{0.00} \leftarrow q_{1.0}\}$,
$C_{36} = \{a_{0.00} \lor b_{1.00} \lor c_{0.00} \leftarrow q_{1.0}\}$.

Finally, the FJRA determines the reduct. Where, there are two-unit clauses in this list, the gene $b$ and $a$ are chosen in turn and all clauses will be satisfied. Later, the set $\{a, b\}$ is returned as the selected genes.

## 4. EXPERIMENTS

In this section, we analyze the proposed technique by implementing it on five tumor datasets. First, we explain a brief description of five benchmark tumor datasets. Next, we present the implementation details, classifier condition, and the experimental results. Finally, we compare the experimental results of proposed technique with the FDM (Jensen and Shen, 2009) and JRA (Jensen et al., 2014) techniques due to utilizing the DM of fuzzy-rough sets.

### 4.1. Tumor datasets

As depicted in Table 3, we have utilized five gene expression benchmark datasets to evaluate the proposed technique. These datasets consist of thousands of real-valued genes. During our experiments, we implemented the proposed technique on these datasets.

**Table 3. Statistics of Tumor datasets.**

| Tumor dataset | No. of genes | No. of samples |
|---|---|---|
| Colon (Alon et al., 1999) | 2000 | 62 |
| Leukemia (Golub et al., 1999) | 7129 | 72 |
| SRBCT (Khan et al., 2001) | 2308 | 83 |
| DLBCL (Rosenwald et al., 2002) | 5470 | 77 |
| Brain_tumor1 (Pomeroy et al., 2002) | 5920 | 90 |

### 4.2. Discernibility power of selected genes

To display the discernibility power of selected genes, we implemented the proposed technique and JRA (Jensen et al., 2014) on three tumor datasets. Then, we demonstrate the distribution of samples in 2D form based on the first two genes selected using JRA and the proposed technique. Herein, the generated plots demonstrate the quality of clusters concerning compactness (i.e., the measure of cohesion of samples into a cluster) and isolation (i.e., the measure of separation among a cluster and others).

Figure 1 depicts the distribution of samples for Colon tumor dataset as bi-class data. The first two genes are chosen by the proposed technique, that can cluster the samples much better than JRA, especially for class annotated by red circles.

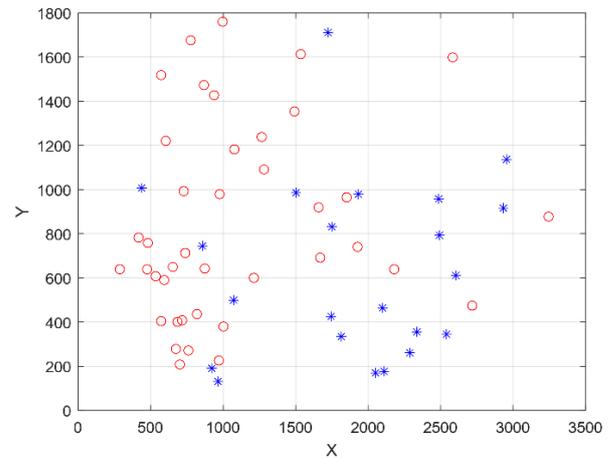

(a)  Genes {66,175} given by JRA

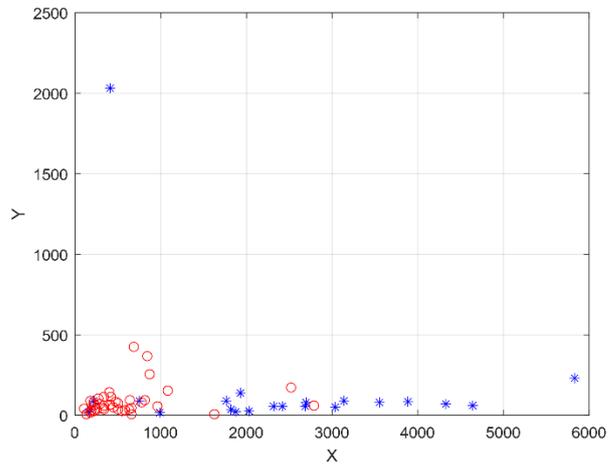

(b) Genes {249,1895} given by proposed technique

Fig. 1. Distribution of Colon samples from prospect of two best genes.

Figure 2 shows the distribution of samples on the Leukemia tumor dataset. Also, in this case, the compactness and isolation of two classes, from the prospect of two best genes selected by the proposed technique, is noticeable.

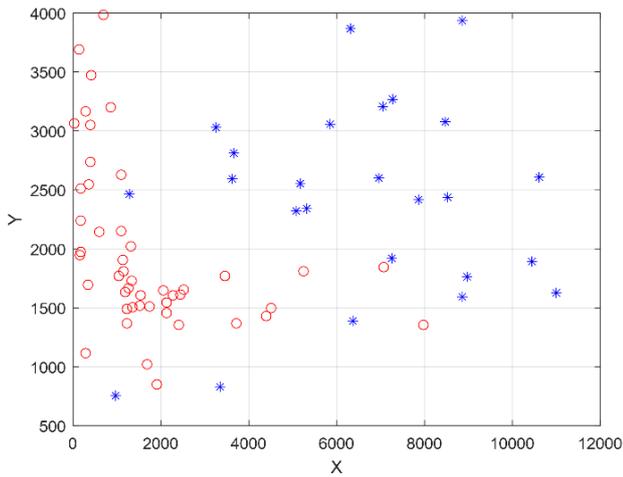

(a) Genes {4196,854} given by JRA

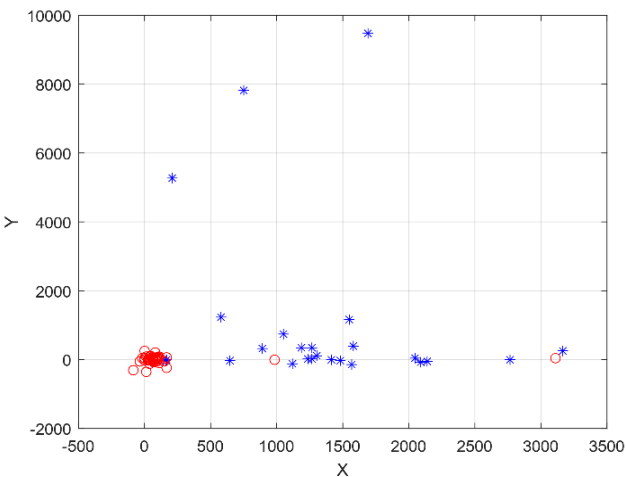

(b) Genes {3252,6277} given by proposed technique

Fig. 2. Distribution of Leukemia samples from prospect of two best genes.

Figure 3 illustrates the distribution of samples on the SRBCT tumor dataset as a multi-class data. In this case, again, the genes selected by the proposed technique can distinguish well the samples of multiple classes regarding compactness (especially for blue class) and isolation (for all four classes).

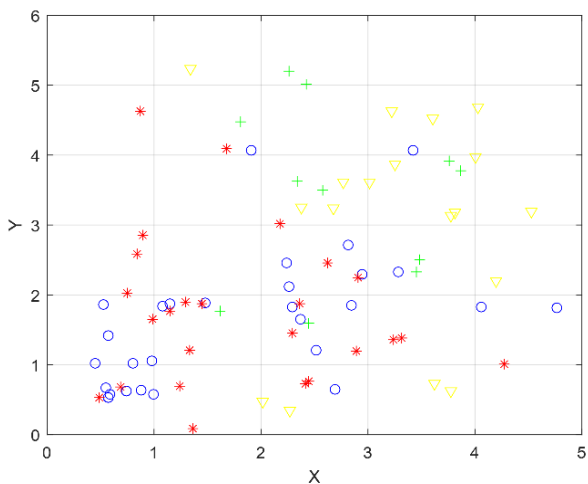

(a) Genes {912,1894} given by JRA

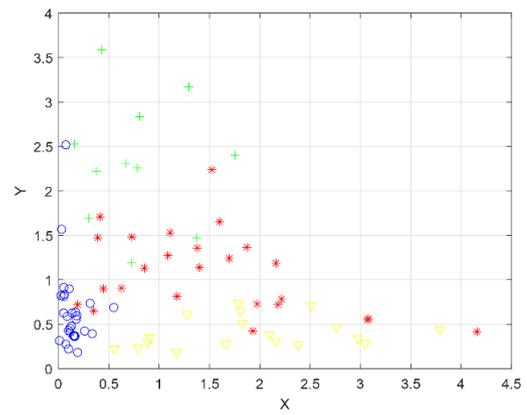

(b) Genes {2050,1158} given by proposed technique

Fig. 3. Distribution of SRBCT samples from prospect of two best genes.

### 4.3. Comparison Results and Discussions

To demonstrate the superiority of proposed technique on the classification of tumor datasets, we compared our experimental results with the FDM (Jensen and Shen, 2009) and JRA (Jensen et al., 2014) algorithms due to utilizing the fuzzy rough sets attribute reduction discernibility matrix. In the JRA, the search method is changed to Johnson reducer algorithm. The decision tree-based classifier, C4.5 (Quinlan, 1996), is used for tumor classification. Also, leave-one-out cross-validation is adopted to evaluate the classification accuracy rate. The fuzzy connectives Lukasiewicz t-norm, $\max(x + y - 1, 0)$, Lukasiewicz t-conorm, $\min(x + y, 1)$ and Lukasiewicz fuzzy implicator, $\min(1 - x + y, 1)$ are used in discernibility functions. Also, the fuzzy relation in equation (4) is employed for evaluating similarity analysis. We implemented all the aforementioned approaches using the Matlab R2014b. First, we examined the effect of selected genes on the classification of tumor datasets. Table 4 summarizes our experimental results of leave-one-out cross-validation while the best accuracy rate for each dataset is highlighted in boldface. According to these accuracies and their averages, the genes selected by the proposed technique are more effective on classification than those chosen by other methods. However, the accuracy rate of JRA is better in the DLBCL dataset.

**Table 4. Classification accuracy of selected genes.**

| Tumor dataset | All genes | FDM | JRA | Proposed Technique |
|---|---|---|---|---|
| Colon | 79.03 | 64.52 | 64.52 | 82.26 |
| Leukemia | 80.56 | 77.78 | 93.05 | 93.05 |
| SRBCT | 80.72 | 66.27 | 48.19 | 81.92 |
| DLBCL | 85.71 | 83.12 | 85.71 | 79.22 |
| Brain_tumor1 | 68.88 | 58.89 | 70.00 | 77.77 |
| Average of accuracy rate | 78.98 | 70.06 | 75.32 | 82.84 |

Table 5 depicts the average number of selected genes from each dataset separately. Since the proposed technique discriminates the objects with the same class labels from those have different labels, the number of satisfied clauses is higher compared to other approaches. Table 5 depicts the number of selected genes.

**Table 5. Average number of selected genes.**

| Tumor dataset | All genes | FDM | JRA | Proposed Technique |
|---|---|---|---|---|
| Colon | 2000 | 4.00 | 4.16 | 6.08 |
| Leukemia | 7129 | 3.00 | 3.00 | 5.90 |
| SRBCT | 2308 | 4.00 | 4.63 | 5.92 |
| DLBCL | 5470 | 3.00 | 3.12 | 5.98 |
| Brain_tumor1 | 5920 | 4.00 | 4.03 | 5.98 |
| Average | | 3.60 | 3.79 | 5.97 |

Table 6 illustrates the computational cost of the proposed technique compared with the FDM and JRA. Due to the number of selected genes for the proposed method is higher than the FDM and JRA on average, it has higher computational complexity. However, the JRA is faster than FDM and our technique.

**Table 6. CPU time of methods in selecting genes.**

| Tumor dataset | FDM | JRA | Proposed technique |
|---|---|---|---|
| Colon | 0.336 | 0.045 | 0.148 |
| Leukemia | 0.996 | 0.170 | 0.614 |
| SRBCT | 0.552 | 0.146 | 0.257 |
| DLBCL | 0.782 | 0.122 | 0.615 |
| Brain_Tumor1 | 1.455 | 0.313 | 0.797 |
| Average of CPU time | 0.824 | 0.159 | 0.486 |

## 4. CONCLUSIONS

In this research, we proposed a new gene selection technique based on the FDM to increase the accuracy of tumor classification. To meet that requirement, we defined a new criterion to measure the fuzzy discernibility of each entry in the discernibility matrix. According to this criterion, we discriminated between the objects with the same class labels and the objects with the different class labels. Also, we described a new fuzzy discernibility function that considers the effectiveness of objects with the same class labels. Our experiments on five real-world tumor datasets demonstrate that the proposed algorithm is appropriate for selecting a compact set of functional genes compared to the-state-of-the-art approaches.

## ACKNOWLEDGMENTS

The authors declare that there is no conflict of interest regarding the publication of this article. Also, they declare that there is no funding for this project.